\documentclass[11pt]{article}
\pdfoutput=1
\usepackage{akbc}
\usepackage{algorithm}
\usepackage{algpseudocode}
\usepackage{algorithm,algpseudocode}
\algrenewcommand\textproc{}
\usepackage{amsmath}
\usepackage{amsfonts}
\usepackage{amssymb}
\usepackage{fancyhdr}
\usepackage{multirow}
\usepackage{booktabs}
\usepackage{graphicx}
\usepackage{wrapfig}  
\usepackage{chngcntr}
\usepackage{float}
\usepackage{pifont}
\usepackage{xcolor}

\usepackage{tikz}
\usetikzlibrary{arrows,automata}

\usepackage{graphicx}

\usepackage{bm}

\ShortHeadings{LinkLogic: A New Method and Benchmark for Explainable Knowledge Graph Predictions}{Kumar-Singh, Polleti, Paliwal, \& Hodos-Nkhereanye}

\finalcopy 

\usepackage[toc,page]{appendix}

\begin{document}

\setlength{\parskip}{0pt}

\title{LinkLogic: A New Method and Benchmark for Explainable Knowledge Graph Predictions}

\author{\name Niraj Kumar-Singh \email nsingh@inboundhealth.com \\
      \addr Inbound Health, Mountain View, CA USA
      \AND
      \name Gustavo Polleti \email gustavo.polleti@usp.br \\
      \addr University of São Paulo, Butanta, São Paulo, Brazil
      \AND
      \name Saee Paliwal \email saeep@nvidia.com \\
      \addr Nvidia, New York, NY USA
      \AND
      \name Rachel Hodos-Nkhereanye \email rachel.hodos@benevolent.ai \\
      \addr BenevolentAI, Brooklyn, NY USA}

\maketitle
\begin{abstract}
While there are a plethora of methods for link prediction in knowledge graphs, state-of-the-art approaches are often black box, obfuscating model reasoning and thereby limiting the ability of users to make informed decisions about model predictions. Recently, methods have emerged to generate prediction explanations for Knowledge Graph Embedding models, a widely-used class of methods for link prediction. The question then becomes, how well do these explanation systems work? To date this has generally been addressed anecdotally, or through time-consuming user research. In this work, we present an in-depth exploration of a simple link prediction explanation method we call LinkLogic, that surfaces and ranks explanatory information used for the prediction. Importantly, we construct the first-ever link prediction explanation benchmark, based on family structures present in the FB13 dataset. We demonstrate the use of this benchmark as a rich evaluation sandbox, probing LinkLogic quantitatively and qualitatively to assess the fidelity, selectivity and relevance of the generated explanations. We hope our work paves the way for more holistic and empirical assessment of knowledge graph prediction explanation methods in the future.  
\end{abstract}

\section{Introduction}
\label{section:intro}

Link prediction in knowledge graphs (KG) is a fairly well-developed field with many important applications, from knowledge base completion \cite{nickel2016review} and product recommendation \cite{He2017} to biomedical hypothesis generation \cite{bonner2021review} \cite{rosalind}. Knowledge Graph Embedding (KGE) models, in which latent representations are learned for entities and relations, are among the most widely-used methods for these link prediction tasks. These embeddings are optimized to be able to reconstruct the semantic relationships present in the original graph, while enabling the prediction of new links. Despite their success, KGE models typically behave as black boxes from the user perspective, hindering their application in domains where high interpretability is needed.

To address these limitations, several ``KGE Explainer'' (KGE-X) methods have recently been proposed to generate explanations for KGE model predictions. However, the field lacks consensus on how to evaluate and benchmark such methods \cite{doshivelez2017towards}. Evaluation to date has typically relied on subjective assessment, via either anecdotal examples \cite{Polleti19, Gusmao2018, Zhang2019} or time-consuming user research \cite{Bianchi20}. The lack of benchmarks has been a limiting factor in the field, preventing rapid iteration on methods development and obscuring any notion of state-of-the-art. 

In the present work, we start by introducing a simple KGE-X method called LinkLogic that builds on the ideas from \cite{Polleti19}. We then describe the construction of the first-ever prediction explanation benchmark, based on familial relationships from the common-sense Freebase \cite{Bollacker} (FB) knowledge graph. Through various experiments, we show how this benchmark can be used as a rich evaluation sandbox for KGE-X methods, revealing both quantitative and qualitative insights into the behavior and performance of LinkLogic. We underscore that this work is not intended to demonstrate that LinkLogic is the state-of-the-art in KGE-X methods. Rather, we provide a deep-dive analysis of explanations generated by this method, aiming to pave the way for more robust evaluation practices of KGE-X methods.

\section{Related Work}
Rule-based systems such as AMIE+ \cite{AMIE} and DRUM \cite{DRUM} represent an alternative class of approaches relative to KGE methods which can mine or learn first-order logic rules to infer missing links in KGs. They are attractive due to the interpretable nature of such rules and their ability to perform inductive reasoning. However, rule-based prediction can be brittle and lack the nuance allowed by richer models. Further, extracted rules describe global as opposed to local reasoning patterns and hence can only describe \textit{how} predictions are generally made, and not \textit{why} a particular link was predicted. There are also a few interesting methods such as pLogicNet \cite{pLogicNet} and RNNLogic \cite{RNNLogic} which adopt a hybrid approach blending the predictive accuracy of KGE methods with the interpretability of rule-based systems. However, while these methods claim to be interpretable, they still only learn global rules and do not provide a way to evaluate the quality of the explanations.  

Path-based methods are the other primary alternative to KGE methods, with recent approaches such as MultiHop \cite{Multi-hop} and DeepPath \cite{DeepPath} using reinforcement learning to train an agent to reason through paths in the KG from query to target node. These methods overcome the limitation of rule-based systems to enable local insight into the model's reasoning for a given prediction. However, they are trained to traverse observed links and hence explanation quality will be highly dependent on the completeness of the underlying KG. Further, they can be fairly challenging and time-consuming to train.  

In contrast to rule-based and path-based systems, KGE-X methods decouple the link prediction model from the explainer model. At the cost of some potential gap in fidelity between the prediction and the explanation, this decoupling enables the two components to be more specialized which could allow for better quality predictions as well as explanations designed to have more desirable properties, e.g. to maximize user comprehension. XKE \cite{Gusmao2018} e.g. consists of an interpretable global surrogate model, a path-based logistic regression, that attempts to mimic the KGE behavior. Despite XKE's ability to produce meaningful explanations, they usually display low fidelity. As an attempt to produce faithful explanations, \cite{Polleti19} what we will refer as KELIX (Knowledge Embedding with Local Interpretable Explanations) employs a local surrogate, logistic regression, trained on binary path-based features extracted from the KGE itself. Despite its methodological advances, KELIX was only evaluated through anecdotal examples and has room for improvement in many dimensions (e.g. feature selection and representation, hyperparameter tuning, etc). LinkLogic directly builds on the ideas presented in \cite{Polleti19} aiming to address some of these limitations. More specifically: (1) explanatory features are extended from binary features to model-informed continuous scores; (2) single hop features are replaced by a more general framework using higher-order paths; (3) the KGE explanation problem is reframed as explaining variation in the KGE score for perturbed queries, solved using regression as opposed to triple classification; and (4) surrogate model coefficients are constrained to be non-negative to improve interpretability.

\section{Methods}
\label{section:methods}

\subsection{Notation and conventions}
\label{section:notation}
A knowledge graph $\mathcal{G}$ consists of a set of entities $e \in \mathcal{E}$, relations $r \in \mathcal{R}$ and ordered triples $(h,r,t) \in \mathcal{T}$ where $h,t \in \mathcal{E}$ correspond to the \textit{head} and \textit{tail} entities linked by some relation $r \in \mathcal{R}$. Let the KGE embedding vectors for $h$ and $t$ be $\textbf{h}$ and $\textbf{t}$, respectively, both in $\mathbb{R}^d$. $\mathbf{I} \in \mathbb{R}^{d \times d}$ is the identity matrix. We will use $c$ and $p$ to refer to the child and parent in the query triple ($c$, \textit{parent}, $p$), and $s$ and $p_2$ for any sibling or co-parent of $c$, respectively. Scalars, vectors and matrices are denoted by lowercase italic, lowercase bold, and uppercase bold lettering, respectively. Statistics when not otherwise specified are reported as mean $\pm$ one standard deviation.

\subsection{Our method: LinkLogic}
\label{section:LinkLogic}
Given a set of trained embeddings from a KGE model and a \textit{query triple} $(h,r,t)$, the goal of LinkLogic is to produce a ranked and scored list of paths that best “explain” the KGE score for the query triple. The method is based on generating local perturbations to the embeddings $\mathbf{h}$ and $\mathbf{t}$ and then training a simple surrogate model to explain the variation in the resulting KGE prediction scores based on variation in scores connecting these perturbed entities to nearby paths, as detailed below. \\

\noindent \textbf{Creating perturbed queries:} Given a query triple $q = (h, r, t)$, $n$ perturbations of the query are generated. A single perturbed query $q_i$ = $(\hat{{h}_i}, {r}, \hat{{t}_i})$ is obtained by adding Gaussian noise to $\textbf{h}$ and $\textbf{t}$, as follows: 
\begin{equation}
\label{eqn:perturb_queries}
    \hat{\textbf{h}_i} = \textbf{h} + \alpha \mathcal{N}(0, \sigma^2_h \mathbf{I}), \quad \hat{\textbf{t}_i} = \textbf{t} + \alpha \mathcal{N}(0, \sigma^2_t \mathbf{I}),
\end{equation}
Here, $\alpha$ is a global hyperparameter that controls the magnitude of the perturbation, while $\sigma_h$ and $\sigma_t$ are computed per query triple, as described in Section \ref{section:variance}. \\\\
 \textbf{Identifying candidate paths:} We define a path $P$ of length $l$ in $\mathcal{G}$ to be an ordered sequence $\{ e_1, r_1, e_2, r_2, \dots, r_l, e_{l+1}\}$ with $e_i \in \mathcal{E}$ and $r_i \in \mathcal{R}$. Note that the corresponding triples $(e_i, r_i, e_{i+1})$ may or may not exist in $\mathcal{T}$. We define the path score $S(P)$ as follows:
\begin{equation}
    S(P) = \frac{1}{l}\sum_{j=1}^{l} -\log(1 - f({e}_j, {r}_j, {e}_{j+1})) \label{eqn:pathscore}
\end{equation}
where $f$ is the KGE decoder-dependent scoring function represented as a plausibility score ranging between 0 and 1. Next, let $\mathcal{P}_l^{a}$ be the set of all paths up to length $l$ that start or end with entity $a$, and let $\mathcal{P}_l^{a,b}$ be the set of all paths up to length $l$ that start and end with $a$ and $b$ (in either order). To generate the candidate list of paths, we compute $S(P)$ for $P \in \mathcal{P}_{l}^{h} \cup \mathcal{P}_{l}^{t} \cup \mathcal{P}_{l+1}^{h, t}$ and take the top $m$ paths with the highest scores. A subset of these paths are selected for the final explanation as described next. \\

\noindent \textbf{Training a local surrogate model:} To compute the feature scores for a single perturbed query $\hat{q} = (\hat{h}, {r}, \hat{t})$, first the $h$, $t$ in \{$\mathcal{P}_{l}^{{h}} \cup \mathcal{P}_{l}^{{t}} \cup \mathcal{P}_{l+1}^{{h}, {t}}$\} are replaced with $\hat{h}$ and $\hat{t}$ respectively and then path scores $S(P)$ are computed for the corresponding set of $m$ paths from $\mathcal{P}_{l}^{\hat{h}} \cup \mathcal{P}_{l}^{\hat{t}} \cup \mathcal{P}_{l+1}^{\hat{h}, \hat{t}}$ using Equation \ref{eqn:pathscore}, resulting in a feature matrix $\textbf{X} \in \mathbb{R}^{\text{n} \text{x} \text{m}}$. The path scores for the perturbed query triple $\textbf{y} \in \mathbb{R}^n$ are used as the dependent variable of the surrogate model. Here, we use a simple Lasso regression model with a non-negative weight constraint, i.e.:
\begin{equation}
\label{eqn:lasso}
     \min_{\beta_j \geq 0} \Big(\sum_{i=1}^{n} \big(y_i - \sum_{j = 1}^{m} x_{ij} \beta_j\big)^2 + \lambda \sum_{j=1}^{m}|\beta_j|\Big)
\end{equation}
\textbf{Generating explanations:} Finally, paths with positive coefficients $\beta_j$ are taken to be the explanation, ranked in descending order by their coefficient values. Pseudocode is provided in Appendix \ref{appendix:pseudocode}.

\subsection{Baseline: Path score heuristic}
\label{section:baseline}

In order to assess whether LinkLogic learns a meaningful ranking of candidate paths, we compared it to a simple baseline, which we will refer to as the path score heuristic, that ranks paths based directly on the path scores $S(P)$ from Equation \ref{eqn:pathscore} in descending order, for $P$ in $\mathcal{P}_l^h \cup \mathcal{P}_l^t \cup \mathcal{P}_{l+1}^{h,t}$. When necessary, a subset of paths was extracted by applying a threshold to the scores.

\section{Experiment Setup}
\label{section:experiment_setup}

\subsection{FB13 and ``FB14'' training data preparation}
\label{section:data_prep}
To create an evaluation sandbox to assess explanation quality, we chose Freebase 13 (FB13) \cite{fb13} which has interpretable entities linked by 13 relations (see Figure \ref{fig:fb_relations} for details) including family relations that can be reasoned upon as described in the next section \footnote{\footnotesize{The dataset was accessed from the OpenKE gituhb repo:\newline \href{https://github.com/thunlp/OpenKE/tree/master/benchmarks/FB13}{https://github.com/thunlp/OpenKE/tree/master/benchmarks/FB13}}}. FB13 contains the following relation categories: Family=\textit{child, parent, spouse}; Location=\textit{location, place-of-birth, place-of-death, nationality}; Other=\textit{cause-of-death, ethnicity, gender, institution, profession, religion}. Triples from FB13 were randomly shuffled into training, validation and test splits with proportions 80/10/10, respectively. Additional processing is described in Section \ref{section:fb13_processing}.  

Among the 13 relations in FB13, there are three familial relations: \textit{parent}, \textit{child} and \textit{spouse}. Sibling relationships, while implicit via entities having shared parents, are not represented explicitly. To investigate the model's usage of sibling relations, we created an alternative data version by adding a new \textit{sibling} relationship for a total of 14 relations, which we will refer to as FB14. Two entities $s_i$ and $s_j$ are defined as siblings if they each have exactly two parents $p_k$ and $p_l$ linked to both $s_i$ and $s_j$. For all such pairs, triples $(s_i, sibling, s_j)$ and $(s_j, sibling, s_i)$ were added for a total of 4130 new triples. These were randomly shuffled with the same 80/10/10 proportions and appended to the training, validation and test sets. 

\subsection{Construction of the \textit{Parents Benchmark} for prediction explanations}
\label{section:benchmark}

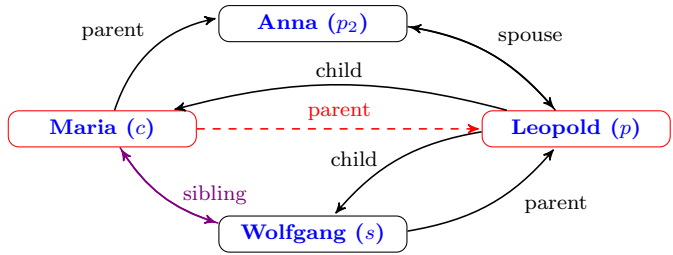
\begin{figure}[h]
\scriptsize
    \centering
    \begin{tikzpicture}[->,>=stealth',shorten >=1pt,auto,node distance=2.6cm,
    semithick,scale=0.7]
\tikzstyle{state}=[fill=red,draw,rectangle,text=white]
\tikzstyle{state2}=[fill=blue,draw,circle,text=white]
\tikzstyle{type1}=[align=center, rectangle, draw, rounded corners, draw=red,
                     thin,bottom color=white, top color=white,
                     text=blue, minimum width=2.5cm]
\tikzstyle{type2}=[rectangle, draw, rounded corners,
                     thin,bottom color=white, top color=white,
                     text=blue, minimum width=2.5cm]

\node[type1]         (A)     at (0.0,0.0)   {\textbf{Maria ($c$)}};
\node[type1]         (B)     at (9.0,0.0) {\textbf{Leopold ($p$)}};
\node[type2]         (C)     at (4,-2.0) {\textbf{Wolfgang ($s$)}};
\node[type2]         (D)     at (4,2.0) {\textbf{Anna ($p_2$)}};

\path[dashed, color=red] (A) edge[bend right=0] node {parent} (B);

\path
(A)  edge[bend right=20, color=violet] node[right=5] {sibling} (C);
\path
(C)  edge[bend left=20,color=violet] (A);
\path
(B)  edge[bend right=20] node[left=5] {child} (C);
\path
(B)  edge[bend right=15] node[above] {child} (A)
;
\path
(C)  edge[bend right=20] node[right=10] {parent} (B);

\path
(A)  edge[bend left] node {parent} (D);

\path
(D)  edge[bend left=20] node {spouse} (B);

\path
(B)  edge[bend right=20] (D);

\end{tikzpicture}
    \caption{\footnotesize{Directed sub-graph representing the Mozart family. Black and red edges are included in FB13. The purple \textit{sibling} edge highlighted was included in FB14. The link in red shows the query triple, (Maria, parent, Leopold).}}
    \label{fig:toy_example_family_graph}
\end{figure}

We constructed a benchmark from the Freebase data that consisted of paths that correctly explain each parent-child link $(c, parent, p)$ according to commonsense reasoning. To illustrate, we start by showing an example family from Freebase revolving around the query triple (Maria Mozart, \textit{parent}, Leopold Mozart). The family (see Figure \ref{fig:toy_example_family_graph}) has two parents, Anna and Leopold, along with their child, Maria and Wolfgang. In reference to the query triple, we use $c$ for Maria, $p$ for Leopold, $p_2$ for Anna, and $s$ for Wolfgang. To explain why the query triple is true, we find natural ``clues'' in the two-hop paths: \{Maria, \textit{parent}, Anna, \textit{spouse}, Leopold\} and \{Maria, \textit{sibling}, Wolfgang, \textit{parent}, Leopold\}. Given that we define siblings based on the sharing of precisely two parents, the latter path deterministically implies that the query triple is true, while the former path can in general be true without the query triple being true. Further, we posit that single hops along these paths are also reasonable components of a good explanation.

\noindent We abstracted this reasoning to construct the Parents Benchmark as shown in Table \ref{table:benchmark}. For each path category, we assigned confidence scores of either 1 or 0.5 indicating whether or not the path deterministically implies the query triple. Additional details are described in Section \ref{section:benchmark_appendix}. The benchmark includes at least one path for 6,260 query triples, with the distribution shown in Figure \ref{fig:benchmark_hist}.

\begin{table}[!h]
\centering
\scriptsize
\begin{tabular}{|c|c|c|}
\hline
\textbf{Path Category}               & \textbf{Confidence} & \textbf{Dataset relevance} \\ \hline
\{$p$, child, $c$\}$^1$                & 1                   & FB13, FB14                 \\ \hline
\{$c$, sibling, $s$, parent, $p$\} & 1                   & FB14 only                  \\ \hline
\{$p$, child, $s$\}                  & 1                   & FB13, FB14                 \\ \hline
\{$c$, sibling, $s$\}                & 1                   & FB14 only                  \\ \hline
\{$c$, parent, $p_2$, spouse, $p$\} & 0.5                 & FB13, FB14                 \\ \hline
\{$c$, parent, $p_2$\}                & 0.5                 & FB13, FB14                 \\ \hline
\{$p$, spouse, $p_2$\}                & 0.5                 & FB13, FB14                 \\ \hline
\end{tabular}
\caption{\footnotesize{The path types included in the Parents benchmark for the query triple $(c, \textrm{parent}, p)$ where $s$ denotes any sibling of child $c$, and $p_2$ corresponds to any co-parent of $c$, i.e. $(c, \textrm{parent}, p_2)$ where $p_2 \neq p$. $^1$Note this is a direct (inverse) representation of the query triple and was only used in Section \ref{section:tautologies}.}}
\label{table:benchmark}
\end{table}

\subsection{KGE model training}
\label{section:training_kge}

The embeddings for FB13 and FB14 were trained using the ComplEx decoder \cite{TrouillonComplEx16} with the Deep Graph Library-Knowledge graph Embeddings (DGL-KE) library \cite{DGL-KE}. We focused on ComplEx due to its ability to learn asymmetric relationships and favorable performance on familial relationships (Supplementary Figure \ref{fig:relation wise MRR}). Hyperparameters are listed in Section \ref{section:kge_hparams}, Supplementary Table \ref{training_hyperparameters}.

\subsection{Evaluation metrics for prediction explanations}
\label{section:evaluation}
We evaluated performance based on three key desirable characteristics \cite{duval2021graphsvx} of prediction explanations: (1) The \textit{fidelity} of a LinkLogic explanation to the KGE model was assessed using the $R^2$ value on a held-out set of target values for the regression from Equation \ref{eqn:lasso}. (2) The \textit{selectivity} was assessed based on the number of paths in an explanation. (3) The \textit{relevance} was assessed using the normalized discounted cumulative gain \cite{jarvelin2002cumulated} among the top \textit{k} paths (NDCG@\textit{k}) based on the coefficients relative to the \textit{Parents Benchmark} confidence scores in Table \ref{table:benchmark}. 

\section{Results}
\label{section:results}

We start with a series of quantitative studies followed by qualitative explorations of family tree configurations and their effects on explanation behavior. 

All LinkLogic explanations were generated using the following hyperparameters: $\alpha=1, \lambda=0.2, n=1000$ and $m=20$ features per relation type and $l = 1$, to extract one-hop and two-hop explanatory paths. With these settings, on a 128 GB CPU, it took less than 10 seconds per query triple to generate LinkLogic prediction explanations. Randomly sampled explanations for all 13 relation types in the FB13 dataset are included in Section \ref{sample_LinkLogic_explanations}. 

\subsection{Analysis of explanations for true and false facts across FB13}
\label{section:true_false}

\begin{figure}[!h]
    \centering
    \includegraphics[width=\textwidth]{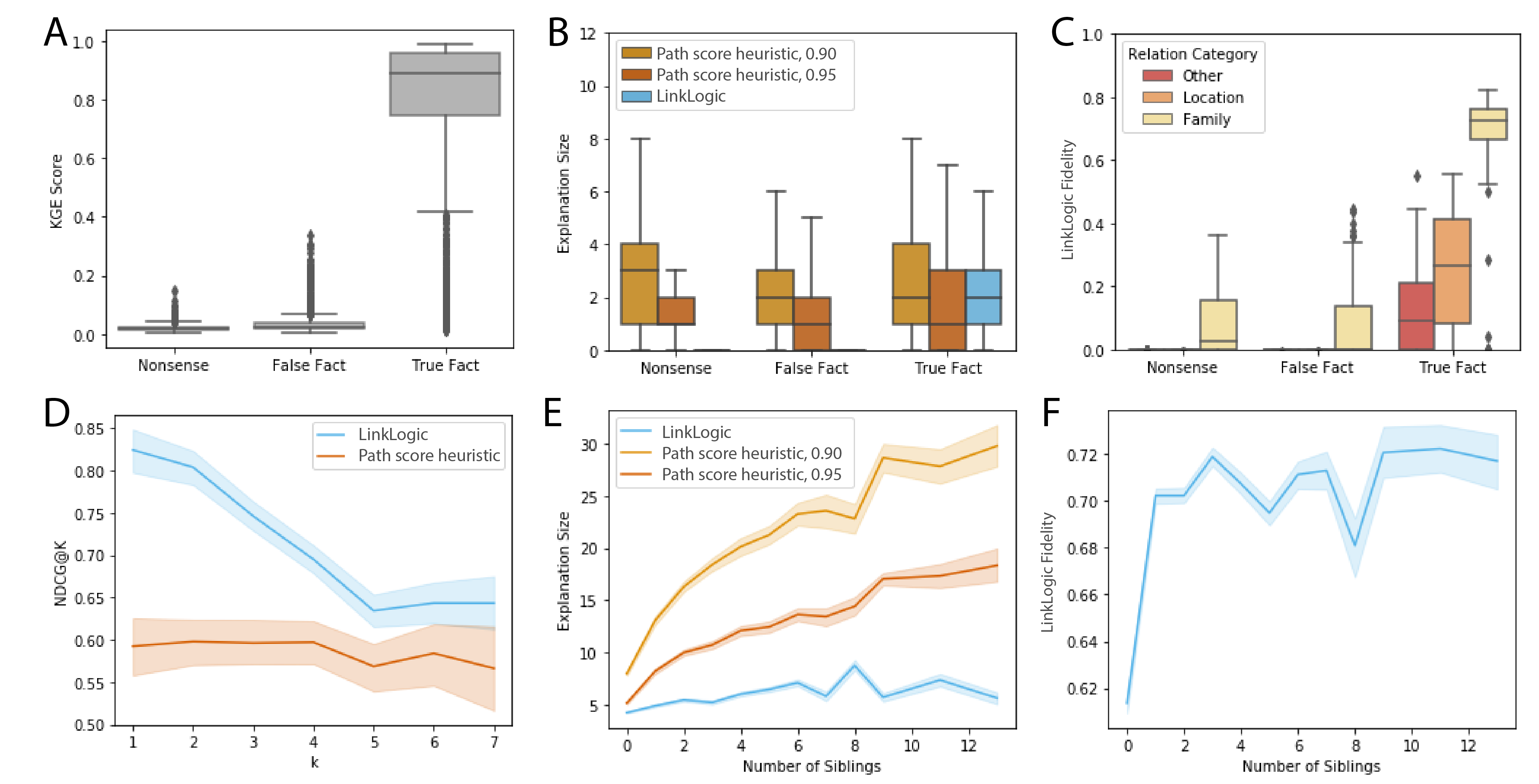}
    \caption{\footnotesize{Quantitative analysis of LinkLogic explanation relevance, selectivity, and fidelity. \textbf{A} KGE prediction scores,  \textbf{B} number of paths per explanation, and \textbf{C} LinkLogic fidelity, for 1300 explanations generated for Nonsense, False, and True query triples across the 3 relation categories in FB13. \textbf{D} Relevance assessment across the full \textit{Parents} benchmark,  \textbf{E} Number of paths per explanation as a function of the number of siblings, and \textbf{F} LinkLogic fidelity versus the number of siblings present in the family. For the path score heuristic, 0.90 and 0.95 correspond to the KGE score threshold at which paths were selected.}}
    \label{fig:quant}
\end{figure}

First, we sought to understand how the fidelity and complexity of the explanations varied throughout FB13, and whether these properties would be different when generated for true vs. false facts. We sampled 100 triples for each of the 13 relation types, for three categories of query triples: \textit{True} facts present in FB13, \textit{False}, i.e. true facts with the tail re-sampled to another entity of the same type, and \textit{Nonsense} triples, which are true triples from FB13 with the tail re-sampled disregarding entity type, e.g. (Paul, \textit{spouse}, New York). Figure \ref{fig:quant}A shows that the KGE scores for \textit{True} facts were significantly higher than those for \textit{False} and \textit{Nonsense} triples (mean of 0.77, 0.03, 0.02, respectively) indicating that the KGE model was trained reasonably well and hence would likely provide a reasonable substrate from which to generate explanations. For each triple, we generated explanations from both LinkLogic and the baseline path score heuristic, thresholding the latter to either 0.90 or 0.95 KGE plausibility scores.

Figure \ref{fig:quant}B shows that while the heuristic path scoring method produced more than one path on average regardless of the triple's truthiness, LinkLogic tended to only produce explanations for \textit{True} triples (mean of 1.9, 0.2, and 0.2 paths for \textit{True}, \textit{False} and \textit{Nonsense}, respectively). Among the LinkLogic explanations, we also quantified the fidelity of each explanation to the KGE scores (see \ref{section:evaluation} for details and Figure \ref{r_squared_plot} for an example). Overall, we observed a trend consistent with Figure \ref{fig:quant}B, where fidelity was significantly higher for \textit{True} facts compared with those in the \textit{False} and \textit{Nonsense} categories (mean 0.29, 0.01, and 0.02 respectively, $t_{1299}>$35, $p<0.0001$). Figure \ref{fig:quant}C shows the fidelity for the three truth categories, broken into the three categories of relations in FB13: Family, Location and Other. Notably, for \textit{True} triples, we observed the highest fidelity for the Family category, versus the Location and Other category. We hypothesized that this is due to the higher-confidence nature of certain familial relationships. On the other hand, the Location relationships can share information amongst themselves, but this relation tends to be noisy and error-prone. Finally, the remaining relations are relatively independent from one another and hence would often lack paths in the graph from which to explain their existence. Nonetheless, we observe nontrivial explanations across most relation types (see Supplementary Tables \ref{FB13-Spouse} to \ref{FB13-CauseOfDeath}). 

\subsection{Analysis of explanations across the Parents Benchmark}
\label{section:quant_parent}

To assess the relevance of selected paths, we computed NDCG@\textit{k} on explanation coefficients (see Section \ref{section:evaluation} for rationale), varying \textit{k} from 1-7, for all triples in the\textit{ Parents Benchmark}. As shown in Figure \ref{fig:quant}D, LinkLogic explanation relevance outperforms that of the path score heuristic by roughly 40\% at $k=1$. This difference is particularly prominent at small \textit{k}, with the margin decreasing as \textit{k} increases.

Recognizing that the number of explanations in the benchmark depends on the number of siblings that links \textit{c} to \textit{p} in the query (\textit{c}, \textit{parent}, \textit{p}), we investigated the properties of explanations as a function of the number of siblings. We found that the number of explanatory paths selected by LinkLogic remained relatively constant, whereas the path score heuristic, as expected, identified more paths as the number of siblings increased (Figure \ref{fig:quant}E), resulting in highly redundant explanations. We also observed a striking pattern in fidelity (Figure \ref{fig:quant}F): explanation fidelity substantially increased between families with zero and one siblings but plateaued with additional siblings. These results indicate that LinkLogic was selective in identifying paths that sufficiently explain the query triple in the presence of redundant evidence through additional siblings.

\subsection{Qualitative evaluation through family tree experimentation}
\label{section:qualitative}

Next we describe a series of experiments with family tree configurations in order to derive further qualitative insights into the properties of LinkLogic. These experiments have in common the shared task of explaining the query triple ($c$, \textit{parent}, $p$), while differing in either the set of paths available at explanation time, or further, in the set of relations available at training time. Note that in order to properly compare the statistics of selected paths across query triples, we restricted the analysis in this section to 895 query triples corresponding to families in FB13 where the child entity $c$ has a single sibling.

\begin{table}
\scriptsize
\centering
\begin{tabular}{l|r|l}
\hline
 \textbf{Experiment}        &   \textbf{Coefficient} & \textbf{LinkLogic Explanation}                                                                                                   \\
\hline
 \multirow{2}{*}{FB13, Child=True}  &         0.983 & leopold\_mozart $\xrightarrow{\text{child}}$ maria\_mozart                                                          \\
   &         0.534 & maria\_mozart $\xrightarrow{\text{parent}}$ anna\_mozart                                                      \\ \hline
 \multirow{5}{*}{FB13, Child=False} &         0.550  & maria\_mozart $\xrightarrow{\text{parent}}$ anna\_mozart $\xrightarrow{\text{spouse}}$ leopold\_mozart        \\
  &         0.329 & leopold\_mozart $\xrightarrow{\text{child}}$ wolfgang\_mozart                                                    \\
  &         0.312 & maria\_mozart $\xrightarrow{\text{parent}}$ anna\_mozart                                                      \\
  &         0.300   & leopold\_mozart $\xrightarrow{\text{spouse}}$ anna\_mozart $\xrightarrow{\text{child}}$ maria\_mozart         \\
  &         0.061 & leopold\_mozart $\xrightarrow{\text{ethnicity}}$ bavarians                                                             \\
 \hline
  \multirow{5}{*}{FB14, Child=False} &         0.451 & leopold\_mozart $\xrightarrow{\text{child}}$ wolfgang\_mozart                                                    \\
 &         0.423 & maria\_mozart $\xrightarrow{\text{sibling}}$ wolfgang\_mozart $\xrightarrow{\text{parent}}$ leopold\_mozart \\
  &         0.352 & maria\_mozart $\xrightarrow{\text{parent}}$ anna\_mozart $\xrightarrow{\text{spouse}}$ leopold\_mozart        \\
  &         0.172 & maria\_mozart $\xrightarrow{\text{parent}}$ anna\_mozart                                                      \\
 &         0.062 & leopold\_mozart $\xrightarrow{\text{location}}$ augsburg                                                                    \\
\hline
\end{tabular}
	\caption{\footnotesize{LinkLogic explanations generated for the query (Maria Mozart, \textit{parent}, Leopold Mozart). Child=True means that the query-inverse triple is available as a prediction explanation. Child=False indicates that this query-inverse has been removed. FB14 Child=False contains the \textit{sibling} relation and does not have the query-inverse triple.}} 
	\label{tab:sample_LinkLogic_explanation}
\end{table}

\subsubsection{Tautologies: ``I am your mother because you are my child''}\label{section:tautologies}

We provided LinkLogic with the “query-inverse” one-hop path \{\textit{p}, \textit{child}, \textit{c}\} as one of the available explanation features. This path, while using a different relation embedding than the query triple, is nonetheless deterministic by being an exact inverse that is guaranteed to be true. Therefore, as a sanity check, we would expect a reasonable prediction explanation system to identify this feature as highly important in the explanation. Indeed, we observed that the query-inverse triple was almost always the top-ranked path (95\% of the time), with a substantially higher coefficient than any other feature (mean of 0.93 compared with the $2^{nd}$-ranked feature with mean of 0.38, $t_{(5157)}$=150, p$<$0.001). Prediction explanations also frequently included a $2^{nd}$ path, \{$c$, \textit{parent}, $p_2$\} through the co-parent $p_2$ (Figure \ref{fig:path_reports}A). The explanation for the Mozart query exemplifies these results; see first row of Table \ref{tab:sample_LinkLogic_explanation}.

\subsubsection{Removing the direct link: ``You are your father's daughter''}\label{section:no_child}

\begin{figure}[h]
\centering
\includegraphics[width=0.9\textwidth]{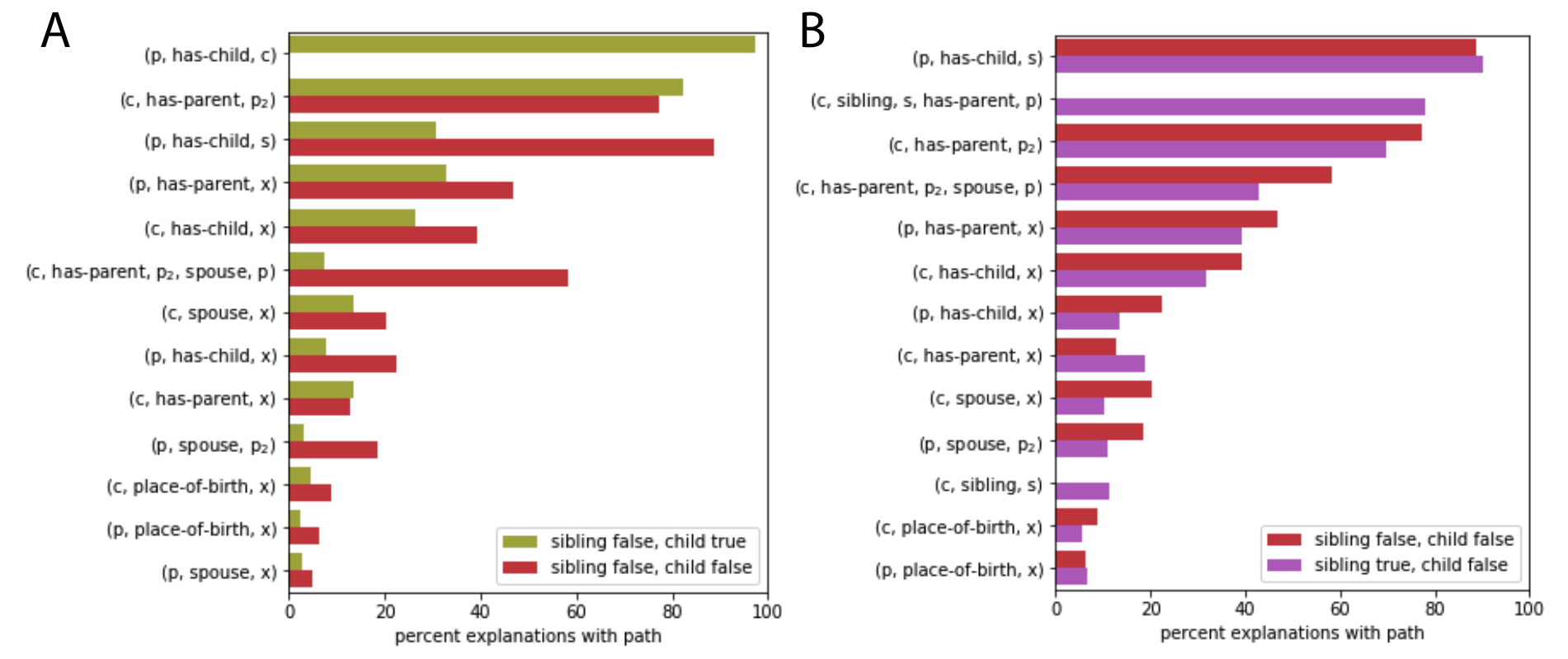} 
\caption{\footnotesize{Most frequently occurring paths in LinkLogic prediction explanations for \textit{parent} relationships, based on various configurations: \textbf{A} before and after removing the direct child link, and \textbf{B} before and after adding the sibling relation to training. $p$=parent, $c$=child, $s$=sibling, $p_2$=co-parent, $x$=any entity that is not $p, c, s, $or $p_2$.}}
\label{fig:path_reports}
\end{figure}

We removed the query-inverse \textit{child} path from the set of features available to the model at explanation time to explore LinkLogic behavior in the absence of this tautology. We continued to allow the model to use other paths involving \textit{child} relations. We observed that the top-ranked path for the Mozart example (second row of Table \ref{tab:sample_LinkLogic_explanation}) is now a two-hop path through the co-parent, Anna, followed by a one-hop path linking the parent Leopold to his son, Wolfgang. Systematic analysis across the 895 query triples reveals that the explanation statistics changed rather substantially relative to the previous configuration, as shown in Figure \ref{fig:path_reports}A. The two most frequent paths align with the top-ranked paths from the Mozart example. Overall we observe a greater variety of explanation paths, most of which are indeed relevant in explaining the query triple. 

\subsubsection{Providing a new vocabulary for siblings: ``Meet your new brother''}
\label{section:siblings}

While we observed a newfound reliance in the explanations on these links from the parent \textit{p} to the sibling \textit{s} (\{\textit{p}, \textit{child}, \textit{s}\} path in Figure \ref{fig:path_reports}A) , the configurations described thus far do not include any (correct) two-hop paths linking \textit{c} and \textit{p} through \textit{s} due to the lack of an explicit \textit{sibling} representation in FB13. Hence, in our next experiment, we asked whether the KGE model identified any relation between true siblings, forcing the creation of an artificial two-hop path linking \textit{c} and \textit{p} through \textit{s}. Re-training the model on FB14, i.e. with explicit \textit{sibling} triples, resulted in sibling-related paths being prioritized in explanations (purple bars, Figure \ref{fig:path_reports}B). Specifically, we saw the reliance of LinkLogic on the correct 2-hop path through the sibling to the parent, \{$c$, \textit{sibling}, $s$, \textit{parent}, $p$\}, as expected. Of note, once the two-hop path through \textit{s} was available to the model, this deterministic path was used even more frequently than the noisier two-hop through $p_2$. These behaviors are reflected in the final row of Table \ref{tab:sample_LinkLogic_explanation} where the two top-ranked paths involve the sibling, Wolfgang, with a lower coefficient for the two-hop path through the co-parent, Anna.

\begin{figure}[h]
\centering
\includegraphics[width=\textwidth]{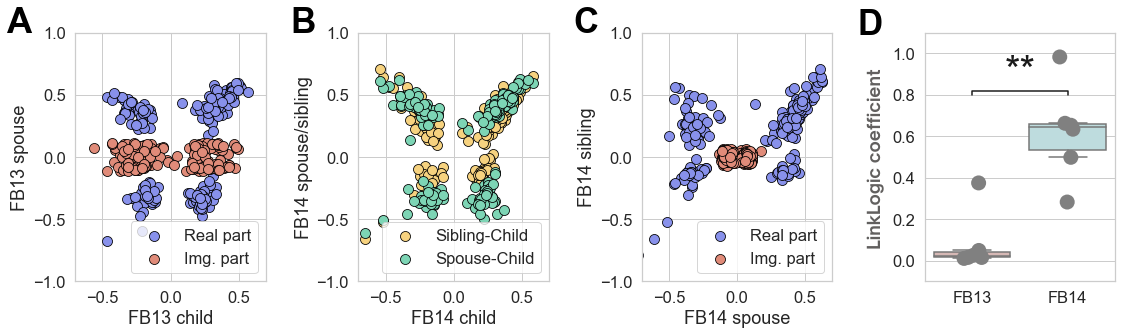}
\caption{\footnotesize{\textbf{A} Scatter plot of real and imaginary parts of the FB13 \textit{spouse} relation embedding against FB13 \textit{children} relation embedding , \textbf{B} Scatter plot of only the real parts of the FB14 spouse (green) and sibling (yellow) relation embedding against the FB14 \textit{children} relation. \textbf{C} Scatter plots of real and imaginary parts of the FB14 sibling embeddings versus the FB14 spouse embeddings. \textbf{D} The change in coefficient of those spouse links (FB13) that were correctly identified as siblings (FB14).}}
\label{fig:pseudo_sib}
\end{figure}

Interestingly, LinkLogic trained on FB13 (without \textit{sibling}), artificially generated \textit{pseudo-sibling} relations using the existing relation \textit{spouse}, which were then used in prediction explanations for 10 query triples. Our hypothesis was that the model generated a stand-in vocabulary to describe the relationship between \textit{c} and sibling \textit{s}, using another symmetric relation (\textit{spouse}). For example, in the explanation for the query triple (Anne Frank, \textit{parent}, Edith Frank Hollander) as shown in Supplementary Figure \ref{fig:anne_frank_family}, the model used explanation (Anne Frank, \textit{spouse}, Margot Frank), creating a \textit{pseudo-sibling} relationship between these true sibling entities. When trained on FB14 (which includes the \textit{sibling} relation), the model correctly assigned 6 of these 10 \textit{pseudo-siblings} to the sibling relationship. 

In the relation embeddings themselves, qualitatively, we see higher covariance between the \textit{spouse}--\textit{sibling} relationship versus the \textit{spouse}--\textit{child} relationship between Figure \ref{fig:pseudo_sib}A and B, and a separation of information between the \textit{sibling-child} relationship and \textit{spouse-child} relationship (yellow and green markers in Figure \ref{fig:pseudo_sib}B). The dispersion of the imaginary part of the \textit{child} embeddings compared to the \textit{spouse} embedding in \ref{fig:pseudo_sib}A shows that the imaginary part of the complex embeddings successfully differentiates between those relationships that are symmetric (e.g. spouse) versus asymmetric relationships (e.g. child). For the \textit{sibling}--\textit{spouse} relationship shown in Figure \ref{fig:pseudo_sib}C, we see a zeroing of the \textit{spouse} and \textit{sibling} ComplEx embedding component, as both relations are symmetric. Finally, in Figure \ref{fig:pseudo_sib}D, we see a significant difference in coefficient between the FB13 \textit{pseudo-sibling} edges and the correctly-assigned FB14 \textit{sibling} edges (U=1.0, p=0.004).

\section{Discussion}
\label{section:discussion}

In this work, we introduced a method for explaining KGE link predictions that builds on an existing method KELIX, while providing various practical modifications. Using family structures in FB13, we also developed a novel evaluation framework and benchmark for rigorously testing KGE-X methods. Applying this framework, we were able to quantify explanation relevance and show substantial improvements over a heuristic baseline. We observed consistently high fidelity in explaining true facts while observing a “reluctance” of the model to produce explanations for false facts. We also demonstrated the selectivity of LinkLogic explanations even when presented with an over-abundance of options from which to choose, exemplified by families with many siblings.  

Through experimenting with family tree configurations, we observed a number of interesting model behaviors. First, in analyzing explanations before and after removing the $\{p, child, c\}$ path from the feature set, we saw the explanation characteristics shift toward higher-order and more varied paths. The observed behavior suggests a level of flexibility which may be beneficial for adapting this method to user-facing products. We also explored the usage of sibling entities and relationships. The high prevalence of paths to sibling entities even without the \textit{sibling} relationship (in FB13 dataset) suggests that a level of inference is possible in the LinkLogic explanations. The generation of \textit{pseudo-siblings} that were then correctly identified as \textit{siblings} (as demonstrated using the FB14 dataset) further illustrates the flexibility of the model to use inferred information for prediction explanation.

We note a couple key limitations of this study. First, we focused on a relatively narrow experimental space, studying performance with one decoder trained on one dataset, focusing on one explanation task, and comparing to one baseline. To reiterate, this was an intentional deep-dive as a means to help us understand and reason about model behavior while establishing a novel evaluation framework. In future work we plan to scale this analysis across these additional dimensions in order to further contextualize LinkLogic relative to other methods and evaluate performance on more difficult tasks. Second, our analysis is based on a random split of the triples in the training data, however the familial relationships are each represented by two distinct directed edges, e.g. $(p, spouse, p_2)$ and $(p_2, spouse, p)$. Hence, most familial links will have been represented in at least one direction in the training data, meaning that this study has focused primarily on observed links. While this can be useful for model introspection, we plan to study the usage of inferred links more carefully by creating robust data splits in future. 

We foresee many interesting extensions to LinkLogic. First, while here we only focus on explaining why a link is true, it is also possible to set up the model to explain why a link is \textit{not} true by simply changing the labels in the regression problem from $-log(1-f)$ to $-log(f)$ for small plausibility scores, e.g. $f<0.5$. Similarly, one could extend LinkLogic to leverage such negative examples in prediction explanations by constructing features of the form $-log(f)$ instead of $-log(1-f)$. This would enable explanations akin to statements like ``\textit{I believe the ground will be dry because I know that it is not raining.}'' Finally, LinkLogic is currently restricted to using a limited set of decoders that ensure local smoothness under perturbation. However, this is easily addressed by replacing the linear regression model with a nonlinear alternative, e.g. based on decision trees. 

With the recent emergence of highly performant and flexible large language models, it would be interesting to evaluate such models' abilities to generate meaningful explanations from KGE-based link predictions. This may be possible, e.g. using a retrieval-augmented generation approach. However, such approaches would have no notion of fidelity to the KGE model, and hence would not be useful for introspecting the base model. LinkLogic, on the other hand, is designed to have fidelity to the underlying KGE model and hence can be used not just to generate meaningful explanations, but also to gain insight into the performance and characteristics of the underlying model.

As machine learning systems are increasingly being applied to support high-stakes decision making, methods for explaining black box models are crucial. We hope that our work proves to be an important step toward prying open the black box and enabling more rigorous evaluation of an emerging class of such methods.

\bibliography{LinkLogic}
\bibliographystyle{plainnat}

\section*{Appendix}
\counterwithin{table}{subsection}
\counterwithin{figure}{subsection}
\appendix
\counterwithin{figure}{section}
\section{Methods and experimental setup}
\subsection{FB13 data processing details}
\label{section:fb13_processing}
As described in the main text, edges were split randomly into training, validation, and test sets. The largest connected component was then identified in the training set, and the small fraction of edges not connected to this component were filtered out. Edges from the validation and test sets linked to entities that were no longer represented in the filtered training set were accordingly removed. This resulted in a final count of 276,690, 34,287 and 34,292 edges in the training, validation and test sets, respectively.

Entity types are not explicitly represented in the FB13 data but are implicit based on the types of relationships connecting them. To enable experimentation with entity types, simple heuristics were applied to assign each entity to a single entity type among the following types: Person, Location, Institution, Profession, Ethnicity, CauseOfDeath, Religion, and Gender. See code for details. \href{https://github.com/niraj17singh/LinkLogic}{https://github.com/niraj17singh/LinkLogic}.
\counterwithin{figure}{subsection}
\begin{figure}[h!]
    \centering
    \includegraphics[width=0.6\textwidth]{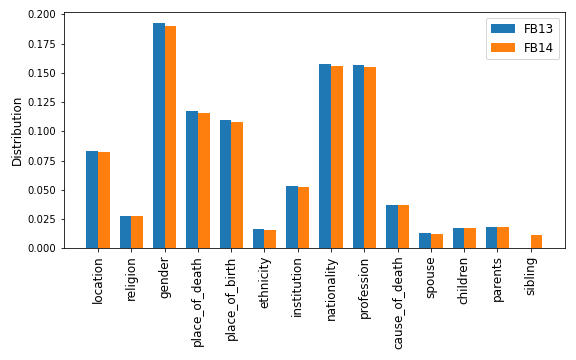}
    \caption{\footnotesize{Distribution of different relation types in the FB13 and FB14 dataset}}
    \label{fig:fb_relations}
\end{figure}

\subsection{Details for construction of prediction explanation benchmark}
\label{section:benchmark_appendix}
For each query triple $(c, parent, p)$ in FB14, individual paths matching each path category were identified and extracted for the benchmark. Due to the bidirectional nature of the familial relationships, e.g. $(h, \textit{spouse}, t) \implies (t, \textit{spouse}, h)$, all paths present in the graph representing the conceptually equivalent information were included in the benchmark. Paths were included in the benchmark regardless of whether the corresponding edges were split in (or across) training, validation and test sets, with these edge splits noted as metadata in the benchmark. 

\counterwithin{figure}{subsection}
\begin{figure}[H]
\centering
\includegraphics[width=0.6\textwidth]{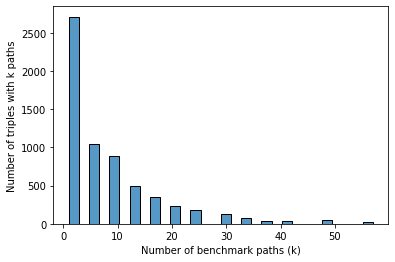}
\caption{\footnotesize{The distribution of benchmark path count per query triple in the Parents Benchmark.}}s
\label{fig:benchmark_hist}
\end{figure}

\subsection{KGE training}

\label{section:kge_hparams}
\begin{figure}[H]
    \centering
    \includegraphics[width=0.6\textwidth]{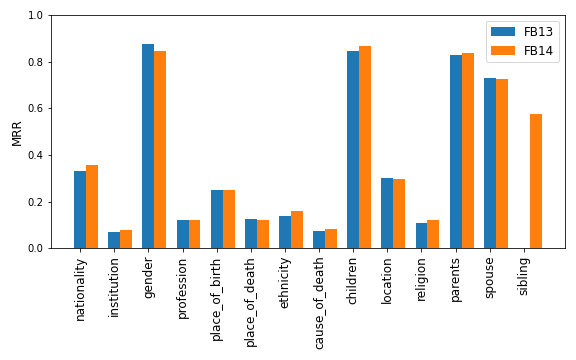}
    \caption{\footnotesize{Comparison of Mean Reciprocal Rank (MRR) for all the relations in the FB13 and FB14 dataset using ComplEx decoder.}
    \label{fig:relation wise MRR}}
\end{figure}

The KGE training was performed using the DGLKE \cite{DGL-KE}, the code publicly available on \href{https://github.com/awslabs/dgl-ke}{github}. The hyperparameters used to train the ComplEx decoder are shown in Table \ref{training_hyperparameters}. Mean reciprocal rank (MRR) per relation on the test set is shown in Figure \ref{fig:relation wise MRR}.

\begin{table}[h]
\centering
\scriptsize
\begin{tabular}{|r|c|}
\hline
\textbf{Hyperparameters} & \textbf{Value} \\ \hline
model\_name               & ComplEx        \\ \hline
batch\_size              & 1000           \\ \hline
neg\_sample\_size        & 200            \\ \hline
hidden\_dim              & 400            \\ \hline
lr                       & 0.1            \\ \hline
max\_step                & 50000          \\ \hline
batch\_size\_eval        & 16             \\ \hline
neg\_adversarial\_sampling & True          \\ \hline
regularization\_coef     & 2e-06          \\ \hline
\end{tabular}
\caption{\footnotesize{Hyperparameters used for training the ComplEx decoder on FB13 and FB14 datasets.}}
\label{training_hyperparameters}
\end{table}

\counterwithin{figure}{subsection}

\section{LinkLogic supplementary information}
\label{section:appendix}

\subsection{Computing LinkLogic noise scaling parameters}
\label{section:variance}

The noise scaling parameters $\sigma_h$ and $\sigma_t$ from Equation \ref{eqn:perturb_queries} are computed at runtime based on the distance of $\mathbf{h}$ and $\mathbf{t}$ to neighboring embeddings. More specifically, neighbors of \textit{h} are found by generating the top $k$ predictions $\{e_1, e_2, \dots, e_k\}$ for the query $(h, r, ?)$ and subsequently finding the top $k$ predictions for each query $(?, r, e_i)$, where $e_i \in \{e_1, e_2, \dots, e_k\}$. This leads to a total of $k^2$ neighbors $\{\tilde{{h}}_1, \tilde{{h}}_2 \dots \tilde{{h}}_{k^2}\}$. Similarly, the neighbors of $t$ can be obtained as $\{\tilde{{t}}_1, \tilde{{t}}_2 \dots \tilde{{t}}_{k^2} \}$. Given the embeddings of each $e \in \mathcal{E}$ as $\textbf{e} \in \mathcal{R}^d$, $\sigma_{h}$ and $\sigma_{t}$ is identified using the following equations:
\begin{align}
    \sigma_h = \sqrt{\frac{\sum_{j=1}^{k^2}\sum_{i=1}^{d} | \tilde{\textbf{h}}_{ji} - \textbf{h}_i|^2}{d k^2}} \label{eqn:head_sigma} \\
    \sigma_t = \sqrt{\frac{\sum_{j=1}^{k^2}\sum_{i=1}^{d} | \tilde{\textbf{t}}_{ji} - \textbf{t}_i|^2}{d k^2}} \label{eqn:tail_sigma}
\end{align}

\subsection{Pseudocode}
\label{appendix:pseudocode}

Algorithm \ref{LinkLogic_pseudocode} provides a pseudocode representation of the LinkLogic method. We denote the KGE decoder as \textit{f} and the trained entity and relation embeddings as \textit{embeddings} as input to the function LinkLogic in the pseudocode. All other inputs are consistent with the naming used in Section \ref{section:methods}.

\begin{algorithm}
   \caption{LinkLogic}
    \begin{algorithmic}[1]
        \Function{LinkLogic}{embeddings, $n$, $l$, $k$, $f$, $\lambda$, $\alpha$, $q=(h, r, t)$} \label{LinkLogic_pseudocode}
        \State $\sigma_{h}, \sigma_{t}$ = \textbf{compute\_sigmas}(embeddings, $q$, $k$)
        \State perturbed\_queries = \textbf{perturb\_queries}(embeddings, $q$, $\sigma_h$, $\sigma_t$, $n$)
        \State paths = \textbf{select\_paths}($q$, $l$)
        \State features = \textbf{compute\_features}(perturbed\_queries, paths, $f$)
        \State labels = \textbf{compute\_labels}(perturbed\_queries, $f$)
        \State coef = \textbf{train\_surrogate\_model}(features, labels, $\lambda$)
        \State explanation = \textbf{extract\_nonzero\_paths}(features, coef)
        \State \Return explanation
        \EndFunction
    \end{algorithmic}
\end{algorithm}

\subsection{Example computation of LinkLogic fidelity}

\label{sample_LinkLogic_explanations}

\begin{figure}[H]
\centering
\includegraphics[width=0.6\textwidth]{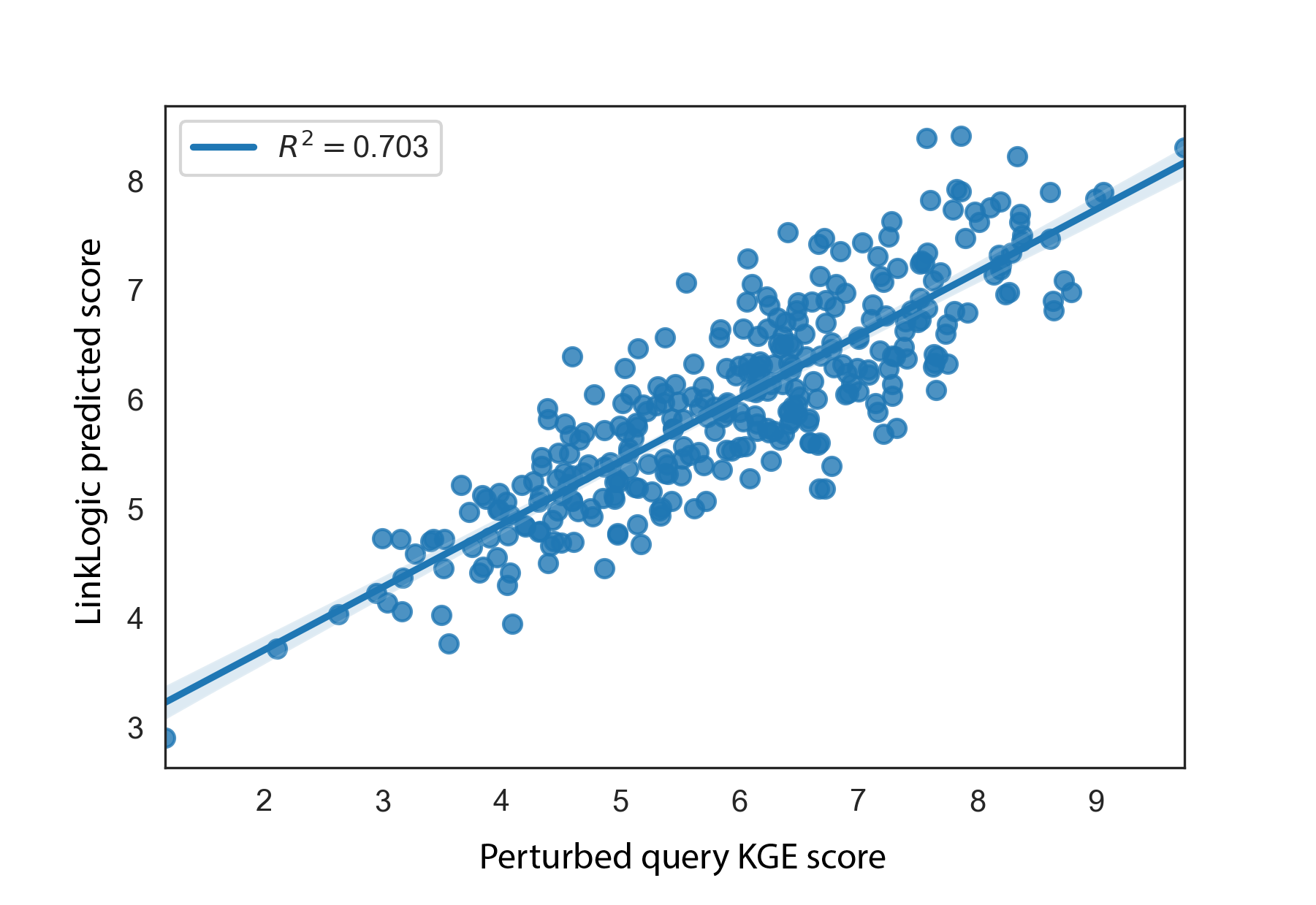}
\caption{\footnotesize{A scatter plot of path scores for 1000 perturbed queries on the $x$-axis and prediction of the surrogate model on $y$-axis.}}
\label{r_squared_plot}
\end{figure}

 For a given query $(\textrm{Maria Mozart}, parent, \textrm{Leopold Mozart})$, Figure \ref{r_squared_plot} shows the path scores for the $n = 1000$ perturbed queries on the x-axis and the predictions from the surrogate model on the y-axis. The fidelity of the LinkLogic explanation is quantified using the $R^2$ score, i.e, the variance explained by the surrogate model. In this example $R^2 = 0.703$.

\counterwithin{figure}{subsection}

\subsection{Sample explanations on FB13 dataset}
The LinkLogic explanations for 5 randomly sampled triples for all the relations in the FB13 dataset (except for \textit{parent}) are shown in Table \ref{FB13-Spouse} to \ref{FB13-CauseOfDeath}. 
\subsubsection{Familial relations}
The LinkLogic explanations for the familial relations: \textit{spouse}, and \textit{child} are shown in Table \ref{FB13-Spouse} and \ref{"FB13-Child"} respectively. We observe that in order to explain \textit{spouse} relation, LinkLogic uses reversed  \textit{spouse} paths, whereas it uses \textit{parents} links to explain \textit{child} links. This indicates that the ComplEx embeddings learn the bidirectional nature of spouse and inverse relation between \textit{parents} and \textit{child}.

\subsubsection{Location relations}
The LinkLogic explanations for the location-based relations: \textit{location}, \textit{place\_of\_birth}, \textit{place\_of\_death}, and \textit{nationality} are shown in Table \ref{FB13-location}, \ref{FB13-place_of_birth}, \ref{FB13-place_of_death} and \ref{FB13-nationality} respectively. We observe that in order to explain the triples with \textit{location} relation, the highest ranked paths use other location-based paths for e.g. using \textit{place\_of\_birth} and \textit{place\_of\_death}.

\subsubsection{Other relations}
The LinkLogic explanations for the other relations: \textit{gender}, \textit{ethnicity}, \textit{religion}, \textit{institution}, \textit{profession} and \textit{cause\_of\_death}, are shown in Table \ref{"FB13-Gender"}, \ref{FB13-ethnicity}, \ref{FB13-religion}, \ref{FB13-institution}, \ref{FB13-profession} and \ref{FB13-CauseOfDeath} respectively. 

\begin{table}[H]
\label{"FB13-Spouse"}
\scriptsize
\centering
\begin{tabular}{|c|c|l|}
\hline
 \textbf{Query}                                                  &  \textbf{Coefficient}  & \textbf{LinkLogic Explanation}                                                                         \\
\hline
 \multirow{2}{*}{moses $\xrightarrow{\text{spouse}}$ zipporah}                                        &         1.341 & zipporah $\xrightarrow{\text{spouse}}$ moses                                                 \\
                                        &         0.138 & moses $\xrightarrow{\text{place\_of\_birth}}$ egypt                                            \\
                                        \hline
                                        
\multirow{2}{*}{ julia\_lennon $\xrightarrow{\text{spouse}}$ freddie\_lennon}                           &         1.21  & freddie\_lennon $\xrightarrow{\text{spouse}}$ julia\_lennon                                    \\
                            &         0.209 & julia\_lennon $\xrightarrow{\text{child}}$ john\_lennon                                        \\
                            \hline
 \multirow{5}{*}{louis\_xii\_of\_france $\xrightarrow{\text{spouse}}$ joan\_of\_france}   &         0.907 & joan\_of\_france $\xrightarrow{\text{spouse}}$ louis\_xii\_of\_france            \\
    &         0.259 & louis\_xii\_of\_france $\xrightarrow{\text{child}}$ renee\_of\_france                             \\
    &         0.165 & joan\_of\_france $\xrightarrow{\text{parent}}$ louis\_xi\_of\_france             \\
    &         0.069 & louis\_xii\_of\_france $\xrightarrow{\text{spouse}}$ mary\_tudor                                 \\
    &         0.04  & louis\_xii\_of\_france $\xrightarrow{\text{place\_of\_birth}}$ chateau\_de\_blois                   \\
    \hline 
 \multirow{7}{*}{ruth\_hussey $\xrightarrow{\text{spouse}}$ bob\_longenecker}                           &         1.05  & bob\_longenecker $\xrightarrow{\text{spouse}}$ ruth\_hussey                                    \\
                            &         0.13  & ruth\_hussey $\xrightarrow{\text{cause\_of\_death}}$ surgical\_complications                         \\
                            &         0.108 & ruth\_hussey $\xrightarrow{\text{place\_of\_birth}}$ providence                                 \\
                            &         0.087 & bob\_longenecker $\xrightarrow{\text{institution}}$ pennsylvania\_state\_university                 \\
                            &         0.058 & ruth\_hussey $\xrightarrow{\text{location}}$ los\_angeles\_county                                   \\
                            &         0.054 & ruth\_hussey $\xrightarrow{\text{location}}$ providence                                           \\
                            &         0.013 & ruth\_hussey $\xrightarrow{\text{place\_of\_birth}}$ los\_angeles\_county                         \\
                            \hline

\multirow{5}{*}{louis\_xviii\_of\_france $\xrightarrow{\text{spouse}}$ marie\_josephine} &         0.497 & marie\_josephine $\xrightarrow{\text{parent}}$ maria\_antonietta\_of\_spain      \\
  &         0.396 & marie\_josephine $\xrightarrow{\text{spouse}}$ louis\_xviii\_of\_france          \\
  &         0.217 & louis\_xviii\_of\_france $\xrightarrow{\text{parent}}$ louis\_dauphin\_de\_france                  \\
  &         0.195 & louis\_xviii\_of\_france $\xrightarrow{\text{parent}}$ marie\_josephe\_of\_saxony                  \\
  &         0.082 & marie\_josephine $\xrightarrow{\text{parent}}$ victor\_amadeus\_iii\_of\_sardinia \\
  \hline

\end{tabular}
\caption{\footnotesize{LinkLogic explanations for five randomly selected examples for the \textit{spouse} relation in the FB13 dataset.}}
\label{FB13-Spouse}
\end{table}

\begin{table}[H]
\scriptsize
\centering
\begin{tabular}{|c|c|l|}
\hline
 \textbf{Query}                                                  &  \textbf{Coefficient}  & \textbf{LinkLogic Explanation}                                                                \\
\hline
 james\_r\_keene $\xrightarrow{\text{child}}$ foxhall\_p\_keene                         &         1.342 & foxhall\_p\_keene $\xrightarrow{\text{parent}}$ james\_r\_keene                         \\
 \hline
 \multirow{3}{*}{ptolemy\_ii\_philadelphus $\xrightarrow{\text{child}}$ lysimachus}                    &         0.828 & lysimachus $\xrightarrow{\text{parent}}$ ptolemy\_ii\_philadelphus                    \\
                     &         0.373 & lysimachus $\xrightarrow{\text{spouse}}$ arsinoe\_ii\_of\_egypt                        \\
                     &         0.077 & ptolemy\_ii\_philadelphus $\xrightarrow{\text{child}}$ arsinoe\_i\_of\_egypt             \\
                     \hline
 \multirow{2}{*}{malietoa\_tanumafili\_i $\xrightarrow{\text{child}}$ malietoa\_tanumafili\_ii} &         0.941 & malietoa\_tanumafili\_ii $\xrightarrow{\text{parent}}$ malietoa\_tanumafili\_i \\
  &         0.746 & malietoa\_tanumafili\_i $\xrightarrow{\text{parent}}$ malietoa\_laupepa                \\
  \hline
 \multirow{2}{*}{robert\_owen $\xrightarrow{\text{child}}$ robert\_dale\_owen}                          &         1.145 & robert\_dale\_owen $\xrightarrow{\text{parent}}$ robert\_owen                          \\
                          &         0.269 & robert\_owen $\xrightarrow{\text{location}}$ powys                                       \\
                          \hline
 \multirow{2}{*}{william\_jennings\_bryan $\xrightarrow{\text{child}}$ ruth\_bryan\_owen}                &         1.24  & ruth\_bryan\_owen $\xrightarrow{\text{parent}}$ william\_jennings\_bryan                \\
               &         0.327 & william\_jennings\_bryan $\xrightarrow{\text{institution}}$ illinois\_college              \\
\hline
\end{tabular}
\caption{\footnotesize{LinkLogic explanations for five randomly selected examples for the \textit{child} relation in the FB13 dataset.}}
\label{"FB13-Child"}
\end{table}

\begin{table}[H]
\scriptsize
\centering
\begin{tabular}{|c|c|l|}
\hline
\textbf{Query}                                                  &  \textbf{Coefficient}  & \textbf{LinkLogic Explanation}                                                           \\
\hline
 richard\_lieber $\xrightarrow{\text{location}}$ indianapolis              &         0.796 & richard\_lieber $\xrightarrow{\text{place\_of\_birth}}$ indianapolis              \\
 \hline
\multirow{2}{*}{ hilary\_p\_jones $\xrightarrow{\text{location}}$ virginia}                  &         0.495 & hilary\_p\_jones $\xrightarrow{\text{place\_of\_birth}}$ hanover\_county            \\
                   &         0.289 & hilary\_p\_jones $\xrightarrow{\text{place\_of\_birth}}$ virginia                  \\
                   \hline
 \multirow{2}{*}{john\_k\_tarbox $\xrightarrow{\text{location}}$ lawrence\_massachusetts}     &         0.663 & john\_k\_tarbox $\xrightarrow{\text{place\_of\_birth}}$ lawrence\_massachusetts     \\

      &         0.279 & john\_k\_tarbox $\xrightarrow{\text{place\_of\_birth}}$ methuen                    \\
      \hline
 \multirow{2}{*}{john\_ball\_1818 $\xrightarrow{\text{location}}$ county\_dublin}             &         0.46  & john\_ball\_1818 $\xrightarrow{\text{place\_of\_birth}}$ county\_dublin             \\
            &         0.319 & john\_ball\_1818 $\xrightarrow{\text{place\_of\_birth}}$ dublin                    \\
            \hline
 \multirow{3}{*}{theodore\_lyman\_1833 $\xrightarrow{\text{location}}$ boston\_massachusetts} &         0.605 & theodore\_lyman\_1833 $\xrightarrow{\text{place\_of\_death}}$ nahant               \\
  &         0.049 & theodore\_lyman\_1833 $\xrightarrow{\text{place\_of\_birth}}$ waltham              \\
  &         0.035 & theodore\_lyman\_1833 $\xrightarrow{\text{place\_of\_birth}}$ boston\_massachusetts \\
  \hline
 john\_ball\_1818 $\xrightarrow{\text{location}}$ county\_dublin             &         0.117 & john\_ball\_1818 $\xrightarrow{\text{institution}}$ st\_marys\_college\_oscott          \\
\hline
\end{tabular}
\caption{\footnotesize{LinkLogic explanations for five randomly selected examples for the \textit{location} relation in the FB13 dataset.}}
\label{FB13-location}
\end{table}

\begin{table}[H]
\scriptsize
\centering
\begin{tabular}{|c|c|l|}
\hline
\textbf{Query}                                                  &  \textbf{Coefficient}  & \textbf{LinkLogic Explanation}                                                      \\
\hline
 gaspara\_stampa $\xrightarrow{\text{place\_of\_birth}}$ padua                     &         1.016 & gaspara\_stampa $\xrightarrow{\text{location}}$ padua                          \\
 \hline
 
  eugene\_viollet-le-duc $\xrightarrow{\text{place\_of\_birth}}$ paris              &         0.233 & eugene\_viollet-le-duc $\xrightarrow{\text{place\_of\_death}}$ lausanne      \\
 \hline
 
 \multirow{4}{*}{r\_m\_hare $\xrightarrow{\text{place\_of\_birth}}$ somerset}                        &         0.838 & r\_m\_hare $\xrightarrow{\text{location}}$ somerset                             \\
                        &         0.206 & r\_m\_hare $\xrightarrow{\text{institution}}$ balliol\_college\_oxford            \\
                        &         0.058 & r\_m\_hare $\xrightarrow{\text{place\_of\_death}}$ oxfordshire                \\
                        &         0.008 & r\_m\_hare $\xrightarrow{\text{institution}}$ rugby\_school                      \\
                        \hline

 \multirow{3}{*}{george\_law\_curry $\xrightarrow{\text{place\_of\_birth}}$ philadelphia}            &         0.284 & george\_law\_curry $\xrightarrow{\text{profession}}$ newspaper              \\
             &         0.167 & george\_law\_curry $\xrightarrow{\text{location}}$ pennsylvania                 \\
             &         0.154 & george\_law\_curry $\xrightarrow{\text{place\_of\_death}}$ portland           \\
             \hline
 \multirow{4}{*}{samuel\_gridley\_howe $\xrightarrow{\text{place\_of\_birth}}$ boston\_massachusetts} &         0.402 & samuel\_gridley\_howe $\xrightarrow{\text{child}}$ laura\_e\_richards         \\
  &         0.265 & samuel\_gridley\_howe $\xrightarrow{\text{institution}}$ harvard\_medical\_school \\
  &         0.054 & samuel\_gridley\_howe $\xrightarrow{\text{place\_of\_death}}$ massachusetts   \\
  &         0.039 & samuel\_gridley\_howe $\xrightarrow{\text{spouse}}$ julia\_ward\_howe         \\
\hline
\end{tabular}
\caption{\footnotesize{LinkLogic explanations for five randomly selected examples for the \textit{place\_of\_birth} relation in the FB13 dataset.}}
\label{FB13-place_of_birth}
\end{table}

\begin{table}[H]
\scriptsize
\centering
\begin{tabular}{|c|c|l|}
\hline
\textbf{Query}                                                  &  \textbf{Coefficient}  & \textbf{LinkLogic Explanation}                                                   \\
\hline

\multirow{3}{*}{pope\_john\_i $\xrightarrow{\text{place\_of\_death}}$ ravenna }                  &         0.506 & pope\_john\_i $\xrightarrow{\text{place\_of\_birth}}$ tuscany              \\
                    &         0.434 & pope\_john\_i $\xrightarrow{\text{place\_of\_birth}}$ ravenna              \\
                   &         0.275 & pope\_john\_i $\xrightarrow{\text{location}}$ ravenna                        \\
\hline
georges\_damboise $\xrightarrow{\text{place\_of\_death}}$ lyon                 &         0.137 & georges\_damboise $\xrightarrow{\text{place\_of\_birth}}$ lyon            \\
 \hline
 
 \multirow{3}{*}{giuseppe\_cesare\_abba $\xrightarrow{\text{place\_of\_death}}$ brescia}          &         0.782 & giuseppe\_cesare\_abba $\xrightarrow{\text{location}}$ brescia               \\
           &         0.223 & giuseppe\_cesare\_abba $\xrightarrow{\text{place\_of\_birth}}$ brescia     \\
           &         0.02  & giuseppe\_cesare\_abba $\xrightarrow{\text{nationality}}$ italy          \\
           \hline
 
 \multirow{4}{*}{jackie\_coogan $\xrightarrow{\text{place\_of\_death}}$ santa\_monica\_california} &         0.415 & jackie\_coogan $\xrightarrow{\text{spouse}}$ betty\_grable               \\
  &         0.342 & jackie\_coogan $\xrightarrow{\text{location}}$ santa\_monica\_california      \\
  &         0.023 & jackie\_coogan $\xrightarrow{\text{cause\_of\_death}}$ cardiovascular\_disease \\
  &         0.013 & jackie\_coogan $\xrightarrow{\text{place\_of\_birth}}$ los\_angeles        \\
   \hline
   \multirow{2}{*}{eddie\_laughton $\xrightarrow{\text{place\_of\_death}}$ hollywood}              &         0.241 & eddie\_laughton $\xrightarrow{\text{place\_of\_birth}}$ sheffield         \\
               &         0.072 & eddie\_laughton $\xrightarrow{\text{cause\_of\_death}}$ pneumonia             \\
               \hline

\end{tabular}
\caption{\footnotesize{LinkLogic explanations for five randomly selected examples for the \textit{place\_of\_death} relation in the FB13 dataset.}}
\label{FB13-place_of_death}
\end{table}

\begin{table}[H]
\scriptsize
\centering
\begin{tabular}{|c|c|l|}
\hline
 \textbf{Query}                                                  &  \textbf{Coefficient}  & \textbf{LinkLogic Explanation}                                                                  \\
\hline
 nikolai\_erdman $\xrightarrow{\text{nationality}}$ russia                   &         0.318 & nikolai\_erdman $\xrightarrow{\text{location}}$ moscow                                     \\
 \hline
  \multirow{2}{*}{johanna\_senfter $\xrightarrow{\text{nationality}}$ germany}                 &         0.226 & johanna\_senfter $\xrightarrow{\text{place\_of\_death}}$ oppenheim                       \\
                  &         0.069 & johanna\_senfter $\xrightarrow{\text{place\_of\_birth}}$ oppenheim                       \\
                  \hline
                  
  \multirow{2}{*}{emilio\_de\_bono $\xrightarrow{\text{nationality}}$ italy}                    &         0.345 & emilio\_de\_bono $\xrightarrow{\text{place\_of\_birth}}$ cassano\_dadda                    \\
                     &         0.088 & emilio\_de\_bono $\xrightarrow{\text{place\_of\_death}}$ verona                           \\
                     \hline
 \multirow{5}{*}{louis\_of\_spain $\xrightarrow{\text{nationality}}$ spain}                    &         0.229 & louis\_of\_spain $\xrightarrow{\text{parent}}$ philip\_v\_of\_spain                        \\
                     &         0.21  & louis\_of\_spain $\xrightarrow{\text{place\_of\_birth}}$ madrid                           \\
                     &         0.065 & louis\_of\_spain $\xrightarrow{\text{place\_of\_death}}$ madrid                           \\
                     &         0.057 & louis\_of\_spain $\xrightarrow{\text{spouse}}$ louise\_elisabeth\_of\_orleans              \\
                     &         0.026 & louis\_of\_spain $\xrightarrow{\text{parent}}$ maria\_luisa\_of\_savoy                     \\
                     \hline

\multirow{2}{*}{horace-benedict $\xrightarrow{\text{nationality}}$ switzerland} &         0.524 & horace-benedict $\xrightarrow{\text{child}}$ albertine\_necker\_de\_saussure \\
  &         0.116 & horace-benedict $\xrightarrow{\text{place\_of\_death}}$ geneva              \\
\hline
\end{tabular}
\caption{\footnotesize{LinkLogic explanations for five randomly selected examples for the \textit{nationality} relation in the FB13 dataset.}}
\label{FB13-nationality}
\end{table}

\begin{table}[H]
\scriptsize
\centering
\begin{tabular}{|c|c|l|}
\hline
  \textbf{Query}                                                  &  \textbf{Coefficient}  & \textbf{LinkLogic Explanation}                                                 \\
\hline
  maria\_montessori $\xrightarrow{\text{gender}}$ female &         0.095 & maria\_montessori $\xrightarrow{\text{child}}$ mario\_montessori\_sr    \\ \hline
  nance\_oneil $\xrightarrow{\text{gender}}$ female      &         0.057 & nance\_oneil $\xrightarrow{\text{location}}$ englewood                    \\
 \hline
  nicola\_rescigno $\xrightarrow{\text{gender}}$ female  &         0.023 & nicola\_rescigno $\xrightarrow{\text{place\_of\_death}}$ viterbo        \\
 \hline
  \multirow{2}{*}{jayne\_mansfield $\xrightarrow{\text{gender}}$ female}  &         0.033 & jayne\_mansfield $\xrightarrow{\text{profession}}$ nude\_glamour\_model \\
   &         0.03  & jayne\_mansfield $\xrightarrow{\text{place\_of\_birth}}$ bryn\_mawr      \\
 \hline
 winnifred\_quick $\xrightarrow{\text{gender}}$ female  &         0.005 & winnifred\_quick $\xrightarrow{\text{place\_of\_birth}}$ plymouth       \\
\hline
\end{tabular}
	\caption{\footnotesize{LinkLogic explanations for five randomly selected examples for the \textit{gender} relation in the FB13 dataset. We note that there is a strong gender bias in the FB13 data (with roughly 7 times as many males as females) which seems to have caused the KGE model to have a poor representation of the \textit{gender} relation (details not shown).}}
	\label{"FB13-Gender"}
\end{table}

\begin{table}[H]
\scriptsize
\centering
\begin{tabular}{|c|c|l|}
\hline
 \textbf{Query}                                                  &  \textbf{Coefficient}  & \textbf{LinkLogic Explanation}                                                 \\
\hline
 
\multirow{2}{*}{ayya\_khema $\xrightarrow{\text{ethnicity}}$ germans}                    &         0.091 & ayya\_khema $\xrightarrow{\text{place\_of\_birth}}$ berlin              \\
                    &         0.034 & ayya\_khema $\xrightarrow{\text{religion}}$ judaism                  \\
                    \hline
 walter\_winchell $\xrightarrow{\text{ethnicity}}$ jew                   &         0.103 & walter\_winchell $\xrightarrow{\text{place\_of\_death}}$ hibbing        \\
 \hline
 \multirow{2}{*}{julian\_tuwim $\xrightarrow{\text{ethnicity}}$ poles}                    &         0.224 & julian\_tuwim $\xrightarrow{\text{place\_of\_death}}$ zakopane          \\
                     &         0.087 & julian\_tuwim $\xrightarrow{\text{institution}}$ university\_of\_warsaw     \\
                     \hline
\multirow{2}{*}{stephen\_i\_of\_hungary $\xrightarrow{\text{ethnicity}}$ hungarian\_people} &         0.291 & stephen\_i\_of\_hungary $\xrightarrow{\text{place\_of\_birth}}$ esztergom \\
  &         0.212 & stephen\_i\_of\_hungary $\xrightarrow{\text{parent}}$ geza\_of\_hungary   \\
  &         0.031 & stephen\_i\_of\_hungary $\xrightarrow{\text{location}}$ esztergom           \\
\hline
\multirow{3}{*}{keenan\_wynn $\xrightarrow{\text{ethnicity}}$ irish\_american}            &         0.383 & keenan\_wynn $\xrightarrow{\text{parent}}$ ed\_wynn                    \\
             &         0.045 & keenan\_wynn $\xrightarrow{\text{place\_of\_birth}}$ new\_york\_state     \\
            &         0.031 & keenan\_wynn $\xrightarrow{\text{cause\_of\_death}}$ pancreatic\_cancer      \\
            \hline

\end{tabular}
\caption{\footnotesize{LinkLogic explanations for five randomly selected examples for the \textit{ethnicity} relation in the FB13 dataset.}}
\label{FB13-ethnicity}
\end{table}

\begin{table}[H]
\scriptsize
\centering
\begin{tabular}{|c|c|l|}
\hline
 \textbf{Query}                                                  &  \textbf{Coefficient}  & \textbf{LinkLogic Explanation}                                                                    \\
\hline
 \multirow{3}{*}{ali\_ibn\_hussayn $\xrightarrow{\text{religion}}$ shia\_islam}               &         0.309 & ali\_ibn\_hussayn $\xrightarrow{\text{child}}$ muhammad\_al-baqir                          \\
               &         0.163 & ali\_ibn\_hussayn $\xrightarrow{\text{parent}}$ husayn\_ibn\_ali                            \\
               &         0.111 & ali\_ibn\_hussayn $\xrightarrow{\text{religion}}$ islam                                  \\ 
               \hline
 alexander\_heriot $\xrightarrow{\text{religion}}$ anglicanism &         0.18  & alexander\_heriot $\xrightarrow{\text{institution}}$ wadham\_college\_oxford       \\ 
 \hline
 catherine\_mcauley $\xrightarrow{\text{religion}}$ roman\_catholic\_church  &         0.129 & catherine\_mcauley $\xrightarrow{\text{location}}$ county\_dublin                             \\
 \hline
 \multirow{3}{*}{charles\_iv\_of\_spain $\xrightarrow{\text{religion}}$ catholicism}          &         0.152 & charles\_iv\_of\_spain $\xrightarrow{\text{child}}$ maria\_louisa \\
           &         0.069 & charles\_iv\_of\_spain $\xrightarrow{\text{place\_of\_birth}}$ portici                       \\
           &         0.037 & charles\_iv\_of\_spain $\xrightarrow{\text{child}}$ ferdinand\_vii\_of\_spain                 \\
           \hline
 joseph\_mccarthy $\xrightarrow{\text{religion}}$ roman\_catholic\_church    &         0.31  & joseph\_mccarthy $\xrightarrow{\text{location}}$ appleton                                    \\
\hline
\end{tabular}
\caption{\footnotesize{LinkLogic explanations for five randomly selected examples for the \textit{religion} relation in the FB13 dataset.}}
\label{FB13-religion}
\end{table}

\begin{table}[H]
\scriptsize
\centering
\begin{tabular}{|c|c|l|}
\hline
 \textbf{Query}                                                  &  \textbf{Coefficient}  & \textbf{LinkLogic Explanation}                                                               \\
\hline
 \multirow{3}{*}{john\_l\_sullivan\_1899 $\xrightarrow{\text{institution}}$ dartmouth\_college}      &         0.225 & john\_l\_sullivan\_1899 $\xrightarrow{\text{place\_of\_death}}$ exeter\_new\_hampshire    \\
       &         0.22  & john\_l\_sullivan\_1899 $\xrightarrow{\text{place\_of\_birth}}$ manchester              \\
       &         0.042 & john\_l\_sullivan\_1899 $\xrightarrow{\text{location}}$ manchester                        \\
\hline
 \multirow{4}{*}{ludwig\_friedlander $\xrightarrow{\text{institution}}$ university\_of\_konigsberg} &         0.307 & ludwig\_friedlander $\xrightarrow{\text{place\_of\_birth}}$ konigsberg                \\
  &         0.25  & ludwig\_friedlander $\xrightarrow{\text{institution}}$ humboldt\_university\_of\_berlin    \\
  &         0.153 & ludwig\_friedlander $\xrightarrow{\text{place\_of\_death}}$ strasbourg                \\
  &         0.11  & ludwig\_friedlander $\xrightarrow{\text{institution}}$ university\_of\_leipzig            \\
  \hline
 leonard\_carmichael $\xrightarrow{\text{institution}}$ harvard\_university       &         0.341 & leonard\_carmichael $\xrightarrow{\text{institution}}$ tufts\_university                 \\
 \hline
 \multirow{4}{*}{kate\_fleming $\xrightarrow{\text{institution}}$ college\_of\_william\_and\_mary}    &         0.271 & kate\_fleming $\xrightarrow{\text{place\_of\_birth}}$ arlington                       \\
     &         0.26  & kate\_fleming $\xrightarrow{\text{cause\_of\_death}}$ drowning                            \\
     &         0.205 & kate\_fleming $\xrightarrow{\text{place\_of\_death}}$ seattle                         \\
     &         0.1   & kate\_fleming $\xrightarrow{\text{location}}$ washington\_united\_states                  \\
     \hline

 \multirow{2}{*}{edward\_mills\_purcell $\xrightarrow{\text{institution}}$ harvard\_university}     &         0.122 & edward\_mills\_purcell $\xrightarrow{\text{institution}}$ purdue\_university              \\
      &         0.076 & edward\_mills\_purcell $\xrightarrow{\text{place\_of\_death}}$ cambridge\_massachusetts \\
      \hline

\end{tabular}
\caption{\footnotesize{LinkLogic explanations for five randomly selected examples for the \textit{institution} relation in the FB13 dataset.}}
\label{FB13-institution}
\end{table}

\begin{table}[H]
\scriptsize
\centering
\begin{tabular}{|c|c|l|}
\hline
\textbf{Query}                                                  &  \textbf{Coefficient}  & \textbf{LinkLogic Explanation}                                                               \\
\hline
\multirow{3}{*}{jeff\_porcaro $\xrightarrow{\text{profession}}$ session\_musician}             &         0.347 & jeff\_porcaro $\xrightarrow{\text{profession}}$ drummer                             \\
             &         0.164 & jeff\_porcaro $\xrightarrow{\text{location}}$ hartford\_connecticut                      \\
             &         0.122 & jeff\_porcaro $\xrightarrow{\text{profession}}$ record\_producer                     \\
             \hline
pavel\_sergeevich\_alexandrov $\xrightarrow{\text{profession}}$ mathematician &         0.328 & pavel\_sergeevich\_alexandrov $\xrightarrow{\text{institution}}$ moscow\_state\_university \\
\hline
 cyril\_gordon\_martin $\xrightarrow{\text{profession}}$ engineer              &         0.253 & cyril\_gordon\_martin $\xrightarrow{\text{place\_of\_death}}$ woolwich                 \\
 \hline
 george\_jessel $\xrightarrow{\text{profession}}$ actor                       &         0.133 & george\_jessel $\xrightarrow{\text{spouse}}$ lois\_andrews                           \\
 \hline
 
 \multirow{4}{*}{yolanda\_king $\xrightarrow{\text{profession}}$ actor}                        &         0.109 & yolanda\_king $\xrightarrow{\text{profession}}$ lgbt\_social\_movements               \\
                         &         0.073 & yolanda\_king $\xrightarrow{\text{place\_of\_birth}}$ montgomery                      \\
                         &         0.053 & yolanda\_king $\xrightarrow{\text{parent}}$ coretta\_scott\_king                      \\
                         &         0.037 & yolanda\_king $\xrightarrow{\text{place\_of\_death}}$ santa\_monica\_california         \\
                         \hline
 
\end{tabular}
\caption{\footnotesize{LinkLogic explanations for five randomly selected examples for the \textit{profession} relation in the FB13 dataset.}}
\label{FB13-profession}
\end{table}

\begin{table}[H]

\scriptsize
\centering
\begin{tabular}{|c|c|l|}
\hline
 \textbf{Query}                                                  &  \textbf{Coefficient}  & \textbf{LinkLogic Explanation}                                                        \\
\hline
 dick\_turpin $\xrightarrow{\text{cause\_of\_death}}$ capital\_punishment      &         0.516 & dick\_turpin $\xrightarrow{\text{cause\_of\_death}}$ hanging                       \\
 \hline
  \multirow{2}{*}{john\_f\_collins $\xrightarrow{\text{cause\_of\_death}}$ pneumonia}            &         0.161 & john\_f\_collins $\xrightarrow{\text{location}}$ boston\_massachusetts             \\
           &         0.059 & john\_f\_collins $\xrightarrow{\text{profession}}$ mayor                      \\
\hline
 kazuo\_ohno $\xrightarrow{\text{cause\_of\_death}}$ respiratory\_failure      &         0.427 & kazuo\_ohno $\xrightarrow{\text{place\_of\_death}}$ yokohama                   \\
 \hline
 \multirow{3}{*}{robert\_mcgehee $\xrightarrow{\text{cause\_of\_death}}$ stroke}               &         0.155 & robert\_mcgehee $\xrightarrow{\text{institution}}$ university\_of\_texas\_at\_austin \\
                &         0.085 & robert\_mcgehee $\xrightarrow{\text{location}}$ mississippi                      \\
                &         0.006 & robert\_mcgehee $\xrightarrow{\text{place\_of\_birth}}$ mississippi            \\
                \hline
 \multirow{2}{*}{robert\_ludlum $\xrightarrow{\text{cause\_of\_death}}$ myocardial\_infarction} &         0.35  & robert\_ludlum $\xrightarrow{\text{place\_of\_death}}$ naples\_florida          \\
  &         0.04  & robert\_ludlum $\xrightarrow{\text{institution}}$ wesleyan\_university            \\
  \hline

\end{tabular}
\caption{\footnotesize{LinkLogic explanations for five randomly selected examples for the \textit{cause\_of\_death} relation in the FB13 dataset.}}
\label{FB13-CauseOfDeath}
\end{table}

\subsection{Anne Frank family}
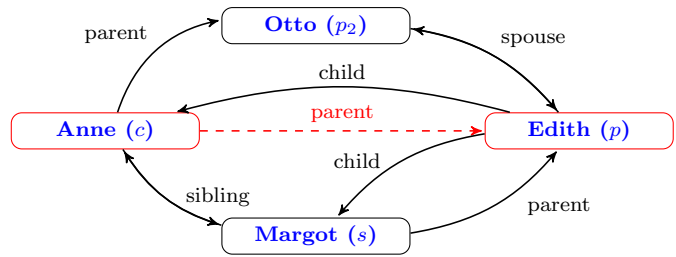
\begin{figure}[H]
\scriptsize
    \centering
    \begin{tikzpicture}[->,>=stealth',shorten >=1pt,auto,node distance=2.6cm,
    semithick,scale=0.7]
\tikzstyle{state}=[fill=red,draw,rectangle,text=white]
\tikzstyle{state2}=[fill=blue,draw,circle,text=white]
\tikzstyle{type1}=[align=center, rectangle, draw, rounded corners, draw=red,
                     thin,bottom color=white, top color=white,
                     text=blue, minimum width=2.5cm]
\tikzstyle{type2}=[rectangle, draw, rounded corners,
                     thin,bottom color=white, top color=white,
                     text=blue, minimum width=2.5cm]

\node[type1]         (A)     at (0.0,0.0)   {\textbf{Anne ($c$)}};
\node[type1]         (B)     at (9.0,0.0) {\textbf{Edith ($p$)}};
\node[type2]         (C)     at (4,-2.0) {\textbf{Margot ($s$)}};
\node[type2]         (D)     at (4,2.0) {\textbf{Otto ($p_2$)}};

\path[dashed, color=red] (A) edge[bend right=0] node {parent} (B);

\path
(A)  edge[bend right=20] node[right=5] {sibling} (C);
\path
(C)  edge[bend left=20] (A);
\path
(B)  edge[bend right=20] node[left=5] {child} (C);
\path
(B)  edge[bend right=15] node[above] {child} (A)
;
\path
(C)  edge[bend right=20] node[right=10] {parent} (B);

\path
(A)  edge[bend left] node {parent} (D);

\path
(D)  edge[bend left=20] node {spouse} (B);

\path
(B)  edge[bend right=20] (D);

\end{tikzpicture}
    \caption{\footnotesize{Directed sub-graph representing the Frank family. The link in red shows the query triple.}}
    \label{fig:anne_frank_family}
\end{figure}

Figure \ref{fig:anne_frank_family} shows the family tree of Anne Frank. Section \ref{section:siblings} uses this as reference to compare the embeddings for \textit{siblings} and \textit{spouse} in the FB14 dataset. 
\counterwithin{figure}{subsection}

\end{document}